\documentclass[10pt,twocolumn,letterpaper]{article}

\usepackage{iccv}
\usepackage{times}
\usepackage{epsfig}
\usepackage{graphicx}
\usepackage{amsmath}
\usepackage{amssymb}

\usepackage{multirow}
\usepackage{mathtools}
\usepackage{textcomp}
\usepackage[accsupp]{axessibility}

\usepackage[pagebackref=true,breaklinks=true,letterpaper=true,colorlinks,bookmarks=false]{hyperref}

\iccvfinalcopy 

\def\iccvPaperID{9} 

\ificcvfinal\fi

\begin{document}

\title{Synthetic Temporal Anomaly Guided End-to-End Video Anomaly Detection}

\author{Marcella Astrid$^{1,2}$, Muhammad Zaigham Zaheer$^{1,2}$, Seung-Ik Lee$^{1,2}$\\
$^{1}$University of Science and Technology, $^{2}$Electronics and Telecommunications Research Institute,\\
Daejeon, South Korea\\
{\tt\small \{marcella.astrid, mzz\}@ust.ac.kr,}
{\tt\small  the\_silee@etri.re.kr}
}

\maketitle

\begin{abstract}
   Due to the limited availability of anomaly examples, video anomaly detection is often seen as one-class classification (OCC) problem.
   A popular way to tackle this problem is by utilizing an autoencoder (AE) trained only on normal data. At test time, the AE is then expected to reconstruct the normal input well while reconstructing the anomalies poorly. 
   However, several studies show that, even with normal data only training, AEs can often start reconstructing anomalies as well which depletes their anomaly detection performance.
   To mitigate this, we propose a temporal pseudo anomaly synthesizer that generates fake-anomalies using only normal data. 
   An AE is then trained to maximize the reconstruction loss on pseudo anomalies while minimizing this loss on normal data. This way, the AE is encouraged to produce distinguishable reconstructions for normal and anomalous frames. 
   Extensive experiments and analysis on three challenging video anomaly datasets demonstrate the effectiveness of our approach to improve the basic AEs in achieving superiority against several existing state-of-the-art models.
\end{abstract}

\section{Introduction}
\label{sec:introduction}

Anomaly detection in video sequences has recently attracted significant attention because of its importance in surveillance systems \cite{sultani2018real,zaheer2020claws,liu2018future,chang2020clustering,li2013anomaly,luo2017revisit,lu2013abnormal,abati2019latent,zhong2019graph}. As anomalous events occur rarely in real-life situations and collecting plenty of anomalous examples can be cumbersome, this task is extremely challenging.
Therefore, anomaly detection is often seen as one-class classification (OCC) problem in which only normal data is used to train a novelty detection model \cite{gong2019memorizing,liu2018future,chang2020clustering,sabokrou2018adversarially,zaheer2020old,liu2008isolation,lee2012anomaly}. 
At test time, the events that do not conform to the learned representations are considered anomalous.

One common way to tackle the OCC problem is by using a deep autoencoder (AE) \cite{hasan2016learning,zhao2017spatio,zaheer2020old,chang2020clustering,luo2017revisit,park2020learning,luo2017remembering,sabokrou2018adversarially,gong2019memorizing}. By training to minimize reconstruction error on normal data, the model is encouraged to extract the features representing normal data in its latent space. This way, at test time, the network is expected to poorly reconstruct the anomalous cases.
However, as previously observed by several researchers \cite{zong2018deep,munawar2017limiting,zaheer2020old}, AEs can also often successfully reconstruct anomalous examples. In such cases, the reconstruction loss between normal and anomalous data may not be discriminative enough to successfully identify the anomalies.

Recently, a new addition to the field of OCC is the idea of utilizing pseudo anomalies generated from the normal training data.
For example, Zaheer \etal \cite{zaheer2020old} fuse two random images from normal data to generate appearance anomalies and use them to train an image classifier. However, this work needs old and new states of the reconstructor and trains in a two-phase scheme. 
Furthermore, the approach is limited to appearance and does not consider any temporal information for anomaly detection.
On the contrary, our work proposes a simple yet highly effective temporal pseudo anomaly synthesizer
to assist the training of an AE in an end-to-end fashion without any bells and whistles. For each pseudo anomaly example as input, our AE model is trained to produce high reconstruction loss. This helps in limiting the capability of AE to reconstruct anomalies at test time. 


We owe the inspiration of our approach to the intuition that detecting fast or suddenly changing motion is significantly important and closely related to detecting anomalies.
For example, it is a common observation that animals often associate strong motion with dangerous situations \cite{evans1993effects}.
The case is similar for humans as well.
For instance, people running unusually may indicate life-threatening situations nearby, such as fires \cite{keating1982myth,elliott1993football} or natural disasters \cite{hu2010study,li2015parameter}.
Moreover, fights or robberies may also be characterized by sudden strong motions.
Some other examples may include riding bikes or vehicles on pedestrian sidewalks, over-speeding vehicles, etc. \cite{aultman1999toronto,sikka2019sharing,muthusamy2015review}.
Therefore, we hypothesize that most of the anomalous events can be characterized by the motion depicted.

\begin{figure}
\begin{center}
\includegraphics[width=\linewidth]{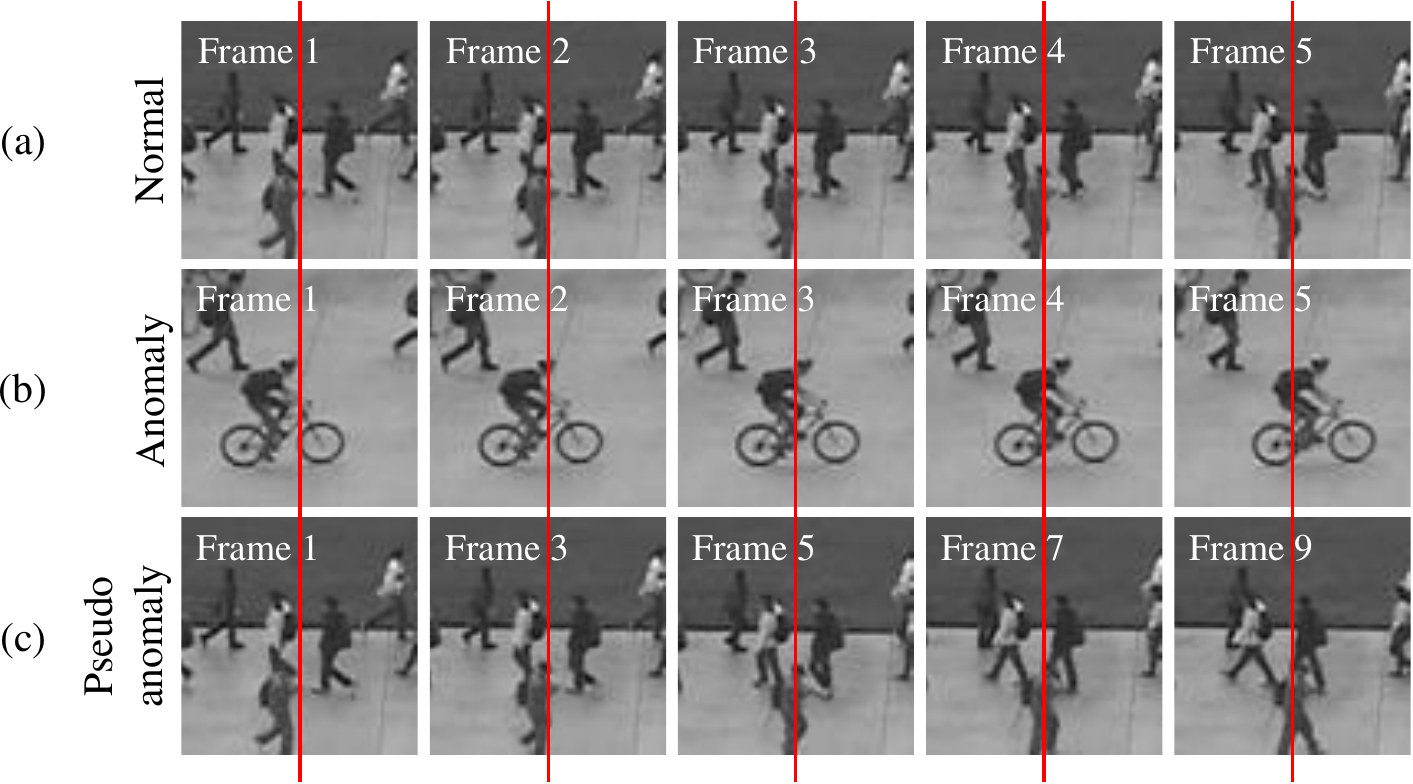}
\end{center}
   \caption{
   Visualization of normal, anomalous, and pseudo anomalous frames. Given red lines as reference, (a) shows a normal movement pattern in which humans are walking in a usual pace. (b) shows an anomalous movement, i.e. the anomalous object almost completely crosses the reference line within few frames. (c) shows the output of our pseudo anomaly synthesizer generated from normal frames which mimics anomalous movements.}
\label{fig:pseudoanomaly}
\end{figure}

To this end, we propose a temporal pseudo anomaly synthesizer that injects synthesized anomaly examples into the training of an AE.
To simulate anomalous movements from normal data, we arbitrarily skip few frames to generate pseudo anomaly sequences as shown in Fig. \ref{fig:pseudoanomaly}. The overall training is then carried out to minimize the reconstruction loss of normal data while maximizing the reconstruction loss of synthesized anomaly data.
Note that, unlike existing motion tracking based anomaly detection methods \cite{basharat2008learning,piciarelli2008trajectory,zhang2009learning,medioni2001event}, our approach does not extract any handpicked motion information. Rather, by complementing a temporal pseudo anomaly synthesizer with a deep AE, we harness the power of deep learning to detect a variety of anomalous activities inside videos. 
Our extensive experiments and analysis demonstrate the superior capability of our approach in three challenging anomaly detection datasets, i.e. Ped2 \cite{li2013anomaly}, Avenue \cite{lu2013abnormal}, and ShanghaiTech \cite{luo2017revisit}.

In summary, the contributions of this paper are as follows:
1) We propose to train one-class classifiers with the assistance of temporal pseudo anomaly synthesizer to produce highly discriminative normal and anomalous reconstructions.
2) Extensive experiments demonstrate the superiority of our method compared to a wide range of existing state-of-the-art (SOTA) works \cite{kim2009observe,mehran2009abnormal,mahadevan2010anomaly,lu2013abnormal,zhang2016video,ramachandra2020street,hasan2016learning,zhao2017spatio,luo2017remembering,luo2017revisit,ravanbakhsh2017abnormal,zaheer2020old,abati2019latent,chang2020clustering,liu2018future,gong2019memorizing,park2020learning,ji2020tam} on three benchmark datasets. 


\begin{figure*}
\begin{center}
\includegraphics[width=\linewidth]{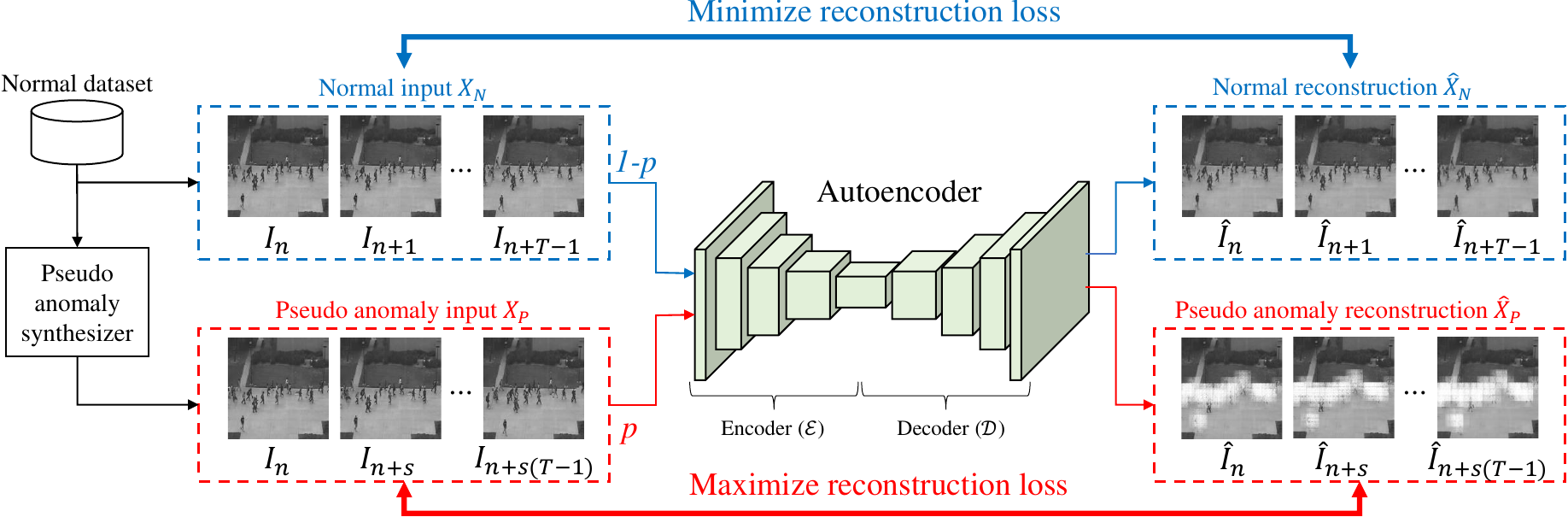}
\end{center}
   \caption{Our approach trains an autoencoder using normal as well as pseudo anomaly sequences. Pseudo anomalies are synthesized using normal data only. The quantity of pseudo anomaly examples is regulated by the probability $p$. The overall configuration is trained to reduce the reconstruction loss for normal inputs while increasing it for pseudo anomaly inputs.}
\label{fig:overall}
\end{figure*}

\section{Related Work}
\label{sec:relatedwork}

Since it is not easy to obtain anomaly examples \cite{patterson2017deep}, a popular approach for anomaly detection is one-class classification (OCC) in which the training is carried out using only normal data. 
Several works train a one-class classifier using the features extracted through object trackers \cite{basharat2008learning,medioni2001event,piciarelli2008trajectory,zhang2009learning}. However, such handpicked features can often limit the generalization capability of the network on different kinds of activities.  

With the recent popularity of deep learning, several researchers \cite{hasan2016learning,zhao2017spatio,zaheer2020old,chang2020clustering,nguyen2019anomaly,luo2017revisit,park2020learning,luo2017remembering,sabokrou2018adversarially,gong2019memorizing} utilize autoencoder (AE) based networks to learn normal data representations.
An AE is trained only on normal data for reconstruction \cite{gong2019memorizing,park2020learning,hasan2016learning,zhao2017spatio,sabokrou2018adversarially,zaheer2020old} or prediction \cite{park2020learning,liu2018future} tasks. At test time, the model is expected to produce poor reconstructions of anomalous data, which correspond to high anomaly scores. Some AE based approaches use appearance information only \cite{sabokrou2018adversarially,zaheer2020old} while several others use both appearance and motion information \cite{zhao2017spatio,chang2020clustering,hasan2016learning,nguyen2019anomaly,luo2017remembering}. 
However, AEs can often reconstruct anomalous data as well \cite{zong2018deep,munawar2017limiting,zaheer2020old}. Consequently, normal and anomalous data become less distinguishable. 

Several attempts have been made to limit the reconstruction capability of an AE on anomalous data. Memory-based networks \cite{gong2019memorizing,park2020learning} employ a memory mechanism over the latent space between the encoder and the decoder of an AE. The network is restricted to use only the memorized normalcy definitions which limits its capability to reconstruct anomalous data.
However, such networks are highly dependent on the memory size and a small-sized memory may also limit their normal data reconstruction capability.
In our approach, we also attempt to limit the reconstruction capability of an AE on anomalous inputs. However, instead of using a memory network, we utilize a pseudo anomaly synthesizer to generate fake anomaly examples and encourage the AE to produce high reconstruction loss on these examples.

The most related to our work is OGNet \cite{zaheer2020old}, which fuses normal images to generate appearance anomalies and train the network using both normal and fake anomaly examples. Moreover, OGNet requires a two-phase adversarial training scheme in which a discriminator is trained based on two previously frozen generator models. On the other hand, our approach of synthesizing anomalies is substantially different. We propose to utilize temporal information rather than appearance to synthesize anomalies. Moreover, our method is end-to-end trainable and complements the conventional training of an AE. 



In order to enhance the discrimination of normal and anomalous data, some researchers \cite{munawar2017limiting, yamanaka2019autoencoding,zaheer2020claws,cleaning2020zaheer,zhong2019graph,zaheer2020self,sultani2018real} propose to deviate from the fundamental definition of OCC by using real anomaly examples during training.
Our method on the other hand utilizes only normal data to synthesize pseudo anomaly examples for training, thus following the conventional OCC protocol. 

\section{Methodology}
\label{sec:methodology}
In this section, we present our proposed STEAL (Synthetic TEmporal AnomaLy guided end-to-end video anomaly detection) network. Due to the unavailability of anomalous examples during training, most of the AE based anomaly detection approaches often fail to discriminate anomalies from normal data at test time. 
Therefore, we propose the utilization of fake anomaly examples, that are generated by our pseudo anomaly synthesizer using only normal training videos, to enhance the performance of AEs.

\subsection{Architecture}

Our overall architecture is shown in Fig. \ref{fig:overall}. We train a conventional AE as our baseline, which takes a sequence of normal frames as input and produces its reconstruction as output. 
To complement the baseline training, we propose a pseudo anomaly synthesizer that generates fake anomaly examples. These examples  are then used for training with a probability $p$.
This way, we limit the reconstruction capability of the AE by forcing it to increase the reconstruction loss on these fake anomaly examples. Finally, the anomaly score is computed using the frame-level reconstruction loss. Each component of our architecture is discussed next:



\subsubsection{Autoencoder}

In order to capture robust representations, autoencoders (AE) are often designed to take multi-frame inputs \cite{gong2019memorizing,park2020learning,hasan2016learning,zhao2017spatio}. Therefore, we set our AE model to take $X$ as input of size $T \times C \times H \times W$, where $T$, $C$, $H$, and $W$ are the number of frames in the input sequence, number of channels, height of frames, and width of frames, respectively. The reconstruction $\hat{X}$ is then given as: 
\begin{equation}
    \hat{X} = \mathcal{D}(\mathcal{E}(X))
    \text{,}
\label{eq:autoencoder}
\end{equation}
where $\mathcal{E}$ and $\mathcal{D}$ are the encoder and the decoder of our model, respectively. 

Conventionally, an AE is trained to reconstruct using only normal data in the training set. At test time, it is expected to well reconstruct the normal data while poorly reconstruct the anomalous examples. However, this is not always the case. AEs can often ``generalize" too well and reconstruct anomalous examples as well \cite{zong2018deep,munawar2017limiting,zaheer2020old}. As there are no anomaly examples utilized in the training time, it can be difficult to train a reconstruction based anomaly detection model \cite{zaheer2020old}. To mitigate this problem, we propose a pseudo anomaly synthesizer that can provide fake anomaly examples. These examples are then used to restrain the generation capability of the AE and encourage it to produce high reconstruction loss on any kind of anomalous inputs.


During training, a sequence of video frames $X$ is given to the network as: 
\begin{equation}
    X= 
\begin{cases}
    X_{N} & \text{with probability $1-p$,}\\
    X_{P} & \text{with probability $p$,}\\
\end{cases}
\label{eq:inputselection}
\end{equation}
where $X_{P}$ is a sequence of frames generated using our proposed pseudo anomaly synthesizer, $X_{N}$ is a sequence of frames from normal training data, and $p$ is the probability that defines the ratio of pseudo anomaly examples used.
Note that the pseudo anomalies are only introduced during training. At test time, we simply input the original sequence of video frames to AE.

\subsubsection{Temporal Pseudo Anomaly Synthesizer}

In our proposed approach, we follow the OCC protocols by utilizing only normal data to carry out the training. Therefore, in order to generate temporal pseudo anomalies, we utilize the normal training videos only. 
Similar to the common practices in training a conventional AE \cite{hasan2016learning,park2020learning}, we extract sequences of frames  $X_{N}$ from a training video $V_i = \{I_1, I_2, ..., I_{K_i} \}$ of length $K_i$ frames by randomly selecting a frame index $n$ from $V_i$ and then taking a fixed number of consecutive $T$ frames thereafter, as described:
\begin{equation}
\begin{multlined}
    X_{N} = (I_n, I_{n+1}, ..., I_{n+T-1}) \\ = (I_{n+t})_{0 \leq t < T, n+T-1 \leq K_i} \text{. }
\label{eq:sequencenormal}
\end{multlined}
\end{equation}

On the other hand, pseudo anomalies $X_{P}$ are synthesized by introducing a skip frame parameter $s$ to Eq. \eqref{eq:sequencenormal} as: 
\begin{equation}
\begin{multlined}
    X_{P} = (I_n, I_{n+s}, ..., I_{n+(T-1)s}) \\ = (I_{n+ts})_{0 \leq t < T, n+(T-1)s \leq K_i, s>1} \text{. }
\label{eq:sequencepseudo}
\end{multlined}
\end{equation}
The skip frame parameter $s$ controls the number of frames we skip to generate temporal pseudo anomaly examples. An example pseudo anomaly sequence with $s=2$ is visualized in 
Fig. \ref{fig:pseudoanomaly}(c).


\subsection{Training}

In order to learn normal representations, a conventional AE is trained on $X_N$ by minimizing reconstruction loss between the input frame $I_{n+t}$ and its reconstruction $\hat{I}_{n+t}$ as follows:
\begin{equation}
    L_N= \frac{1}{T \times C \times H \times W}\sum_{t=0}^{T-1} \left \| \hat{I}_{n+t} - I_{n+t}  \right \|_{F}^{2} \text{,}
\label{eq:aereconloss}
\end{equation}
where $\left \| .  \right \|_{F}$ means Frobenius norm.



For $X_P$ generated by our temporal pseudo anomaly synthesizer, the loss can be similarly defined as:
\begin{equation}
    L_P= - \frac{1}{T \times C \times H \times W} \sum_{t=0}^{T-1} \left \| \hat{I}_{n+ts} - I_{n+ts}  \right \|_{F}^{2}  \text{.}
\label{eq:pseudoreconloss}
\end{equation}
Note the negative sign in Eq. \eqref{eq:pseudoreconloss}, which is introduced to increase the reconstruction loss of pseudo anomaly examples. This helps in limiting the reconstruction capability of our AE on anomalous inputs.

Then, the overall loss $L$ for training takes the form:
\begin{equation}
    L= 
\begin{cases}
    L_N & \text{if } X=X_N \text{,}\\
    L_P & \text{if } X=X_P \text{.}\\
\end{cases}
\label{eq:reconloss}
\end{equation}

\subsection{Anomaly Score}
At test time, concurrent to the existing approaches \cite{gong2019memorizing,park2020learning,liu2018future,hasan2016learning,zhao2017spatio,zaheer2020old,zaheer2020claws}, we predict anomaly scores at frame level. 
Moreover, we compute these scores by utilizing the reconstruction based Peak Signal to Noise Ratio (PSNR). According to Mathieu \etal \cite{mathieu2015deep}, PSNR is often a better assessment of image quality than the reconstruction loss itself. Recently, it has also been utilized in anomaly detection \cite{park2020learning,liu2018future} where PSNR between an input frame and its reconstruction is used to calculate the anomaly score. 
In our approach, we compute PSNR $\mathcal{P}_t$ as: 
\begin{equation}
    \mathcal{P}_t = 10 \text{ log}_{10}  \frac{[M_{\hat{I}_t}]^2}{\frac{1}{R} \left \| \hat{I}_t - I_t  \right \|_{F}^{2} } \text{,}
\label{eq:psnr}
\end{equation}
where $R$ is the total number of pixels in $\hat{I}_t$, $t$ is the frame index, and $M_{\hat{I}_t}$ is the maximum possible value of $\hat{I}_t$. 

Then, following \cite{park2020learning,liu2018future}, we normalize the PSNR value to the range of $[0,1]$ by a min-max normalization over all the frames in a test video $V_i$ as follows:
\begin{equation}
    \mathcal{Q}_t = \frac{\mathcal{P}_t - \min_t(\mathcal{P}_t)}{\max_t(\mathcal{P}_t)-\min_t(\mathcal{P}_t)} \text{,}
\label{eq:minmax}
\end{equation}
where $t$ is the frame index of $V_i$. In Eq. \eqref{eq:minmax}, a higher $Q_t$ represents lower reconstruction loss compared to the other frames in $V_i$ and vice versa. Therefore, we calculate the final anomaly score $\mathcal{A}_t$ as:
\begin{equation}
    \mathcal{A}_t = 1 - \mathcal{Q}_t \text{.}
\label{eq:anomalyscore}
\end{equation}

\section{Experiments}
\label{sec:experiments}

\subsection{Datasets}
We evaluate our approach on three widely popular video anomaly detection datasets, i.e. Ped2 \cite{li2013anomaly}, Avenue \cite{lu2013abnormal}, and ShanghaiTech \cite{luo2017revisit}. We utilize the standard division of the datasets in which training splits consist of only normal videos. Whereas, every video in each of the test sets contains one or more anomalous portions.

\noindent\textbf{Ped2.} This dataset consists of 16 training and 12 test videos \cite{li2013anomaly}.
Pedestrians dominate most of the normal frames whereas anomalies include bikes, carts, or skateboards. 

\noindent\textbf{Avenue.} This dataset contains 16 training and 21 test videos \cite{lu2013abnormal}.
Anomalies include abnormal objects such as bikes and abnormal actions of humans such as unusual walking directions, running, or throwing stuff.

\noindent\textbf{ShanghaiTech.} This is by far the largest one-class anomaly detection dataset \cite{luo2017revisit}. It consists of 330 training and 107 test videos.
The dataset is recorded at 13 different locations having complex lighting conditions and camera angles. In total, the test videos contain 130 anomalous events including running, riding bicycle, and fighting.


\subsection{Experimental Setup}
\noindent\textbf{Evaluation criteria.}
To evaluate our approach, we follow the widely popular frame-level area under the ROC curve (AUC) metric \cite{kim2009observe,mehran2009abnormal,mahadevan2010anomaly,lu2013abnormal,zhang2016video,ramachandra2020street,hasan2016learning,zhao2017spatio,luo2017remembering,luo2017revisit,ravanbakhsh2017abnormal,zaheer2020old,abati2019latent,chang2020clustering,liu2018future,gong2019memorizing,park2020learning,ji2020tam}. The ROC curve is obtained by varying the threshold of anomaly scores to plot false and true positive rates across the whole test set of each dataset. Higher AUC values represent more accurate results. 

\noindent\textbf{Parameters and implementation details.} 
We adopt a generative architecture recently proposed by Gong \etal \cite{gong2019memorizing} as our baseline that takes an input sequence $X$ (Eq. \eqref{eq:autoencoder}) of size $16 \times 1 \times 256 \times 256$ and produces its reconstruction of the same size. All 16 frames are used for computing the reconstruction loss  during training (Eq. \eqref{eq:aereconloss} - \eqref{eq:pseudoreconloss}).
At test time, following \cite{gong2019memorizing}, only 9th frame out of the 16 frames is considered for anomaly score calculation (Eq. \eqref{eq:psnr} - \eqref{eq:anomalyscore}). However, differently from the original implementation, we remove the memory network and utilize only the autoencoder part. Furthermore, we add Tanh output layer to have the output range of $[-1, 1]$. 

The implementation of STEAL Net and the baseline is done in PyTorch \cite{NEURIPS2019_9015}. 
The training is carried out using Adam \cite{kingma2014adam} with a learning rate of $10^{-4}$ and the batch size is set to $4$. The skip frame parameter $s$ in Eq. \eqref{eq:sequencepseudo} is set to $\{2,3,4,5\}$, which means $s$ can be randomly selected as $2$, $3$, $4$, $5$ each time we generate a pseudo anomaly sequence. The probability $p$ for pseudo anomaly in Eq. \eqref{eq:inputselection} is set to $0.01$. 
The baseline in our results refer to the model trained without pseudo anomalies, i.e. the probability $p$ in Eq. \eqref{eq:inputselection} is set to $0$. 

\subsection{Quantitative Results}
\label{sec:SOTA}

Table \ref{tab:sota} shows the AUC comparisons of our proposed STEAL Net on Ped2 \cite{li2013anomaly}, Avenue \cite{lu2013abnormal}, and ShanghaiTech \cite{luo2017revisit} datasets.
Our approach yields better performance than the baseline on all three datasets. Specifically, our approach demonstrates an absolute gain of $5.9\%$, $5.6\%$, and $2.4\%$ AUC on Ped2, Avenue, and ShanghaiTech datasets respectively.  
This shows that our approach successfully improves the capability of the baselines on several diverse datasets.

\begin{table}[]
\caption{AUC performance comparison of our approach with several SOTA methods on Ped2, Avenue (Ave), and ShanghaiTech (Sh). Best performances in each dataset are highlighted as bold whereas second best are highlighted as underlined.}
\resizebox{\linewidth}{!}{
\small
\begin{tabular}{|l|ccc|}
\hline
Methods                  & Ped2 \cite{li2013anomaly}  & Ave \cite{lu2013abnormal} & Sh \cite{luo2017revisit}     \\ \hline
MPPCA \cite{kim2009observe}                        & 69.3\%  & -       & -     \\
   MPPC+SFA \cite{kim2009observe}                     & 61.3\%  & -       & -     \\
   Mehran \etal \cite{mehran2009abnormal}             & 55.6\%  & -       & -     \\
   MDT   \cite{mahadevan2010anomaly}                  & 82.9\%  & -       & -     \\
   Lu \etal \cite{lu2013abnormal}                     & -       & 80.9\%  & -     \\
   LSHF    \cite{zhang2016video}                      & 91.0\%  & -       & -     \\
   Ramachandra and Jones \cite{ramachandra2020street} & 88.3\%  & 72.0\%  & - \\
  
AE-Conv2D \cite{hasan2016learning}           & 85.0\%                                       & 80.0\%                                          & 60.9\%                                            \\
                         AE-Conv3D   \cite{zhao2017spatio}            & 91.2\%                                       & 77.1\%                                          & -                                                 \\
                         AE-ConvLSTM   \cite{luo2017remembering}      & 88.1\%                                       & 77.0\%                                          & -                                                 \\
                         TSC   \cite{luo2017revisit}                  & 91.0\%                                       & 80.6\%                                          & 67.9\%                                            \\
                         StackRNN   \cite{luo2017revisit}             & 92.2\%                                       & 81.7\%                                          & 68.0\%                                            \\
                         AbnormalGAN   \cite{ravanbakhsh2017abnormal} & 93.5\%                                       & -                                               & -                                                 \\
                          OGNet   \cite{zaheer2020old}      & \underline{98.1\%}                                       & -                                          & - \\
                        LSA   \cite{abati2019latent}                 & 95.4\%                                       & -                                               & 72.5\%                                            \\
                         Cluster AE   \cite{chang2020clustering}      & 96.5\%                                       & 86.0\%                                          & \underline{73.3\%}                                            \\
                        Frame-Pred \cite{liu2018future}              & 95.4\%                                       & 85.1\%                                          & 72.8\% \\
                         MemAE   \cite{gong2019memorizing}    & 94.1\%                                       & 83.3\%                                          & 71.2\%                                            \\
                         MNAD-Prediction   \cite{park2020learning}    & 97.0\%                                       & \textbf{88.5\%}                                          & 70.5\%                                            \\
                         MNAD-Reconstruction   \cite{park2020learning}       & 90.2\%                                       & 82.8\%                                          & 69.8\%                                            \\
                         TAM-Net  \cite{ji2020tam}                      & \underline{98.1\%}  & 78.3\%  & - \\
                         \hline
                         Baseline                                                          & 92.5\%                                            & 81.5\%                                               & 71.3\%                                                 \\
                         
                         STEAL Net                                                          & \textbf{98.4\%}                                            & \underline{87.1\%}                                                & \textbf{73.7\%}                                                 \\ \hline

\end{tabular}
}
\label{tab:sota}
\end{table}

Comparing the performance with existing approaches, we achieve superior performances on all three datasets. 
Although MNAD-Prediction \cite{park2020learning} achieves better performance on Avenue compared to our method, its superiority may be attributed to different base architecture as well as training settings, i.e., prediction task. Nevertheless, compared with MNAD-Reconstruction, which has a similar base architecture as ours and performs the same task of reconstruction, our model yields noticeably better performance, thus demonstrating  the  superiority of our approach.

\begin{figure}
\begin{center}
\includegraphics[width=.9\linewidth]{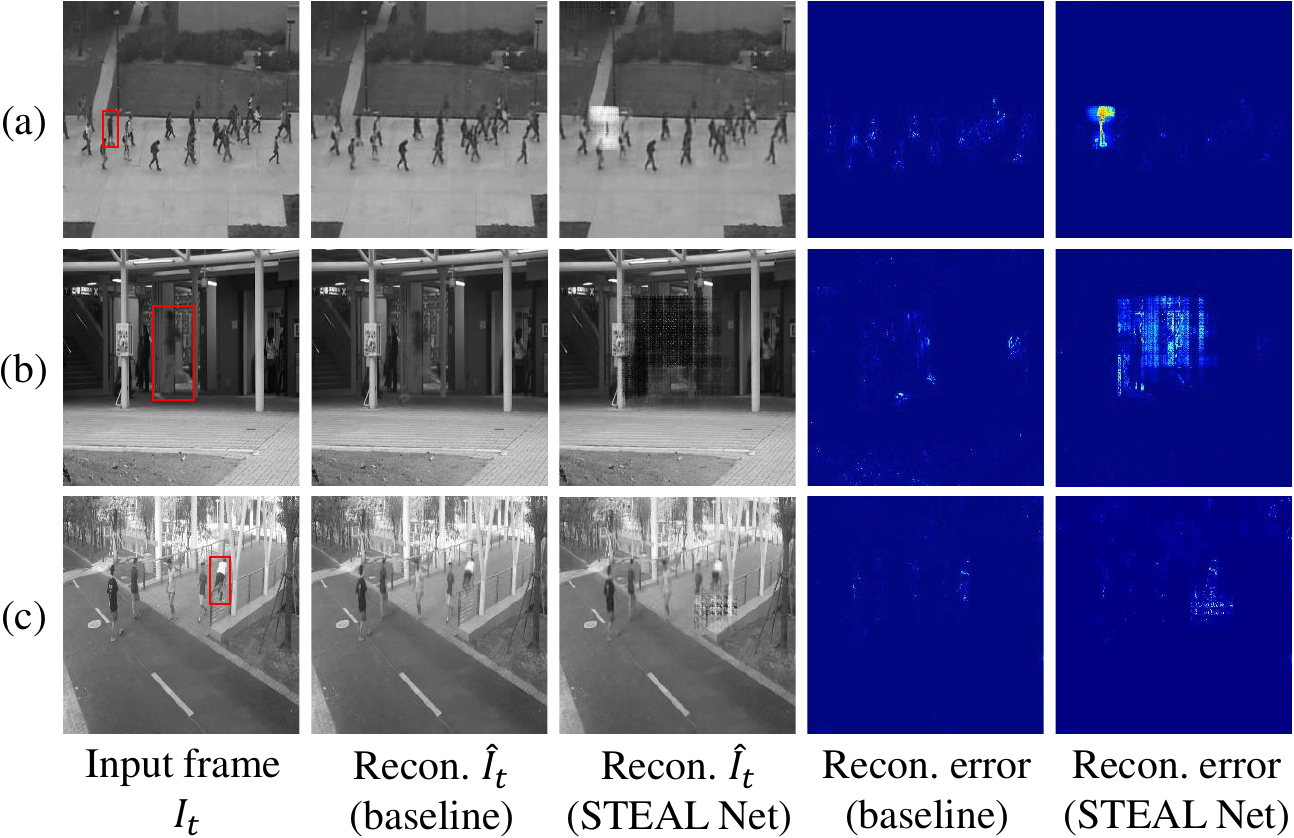}
\end{center}
   \caption{Reconstruction comparison between our approach and the baseline on normal and anomalous frames from Ped2, Avenue, and ShanghaiTech datasets. 
   Our approach specifically attempts to distort the anomalous objects thus producing more discrimination.
   Anomaly ground truths are marked as red boxes.}
\label{fig:qualitativeresults}
\end{figure}

Overall, the superior performance of STEAL Net validates our hypothesis that several different kinds of anomalies may be characterized by the motions present within. Therefore, by utilizing a temporal pseudo anomaly synthesizer, we unleash the potential of a deep autoencoder for more accurate anomaly detection.

\subsection{Qualitative Results}

Input images and the comparisons of their reconstructions produced by STEAL Net and the baseline are shown in the first three columns of Fig. \ref{fig:qualitativeresults}. In addition, reconstruction error heat maps are visualized in the last two columns.
These heat maps are generated by computing the squared error of each pixel between the reconstruction and the input frame followed by min-max normalization. 
As our approach is specifically trained to produce high reconstruction loss on anomalous inputs, it consequently attempts to satisfy this condition by distorting the anomalous regions. On the other hand, concurrent with the reports in several research works \cite{zong2018deep,munawar2017limiting,zaheer2020old}, the baseline reconstructs the anomalies well, thus often failing to produce noticeable discrimination. 



\section{Conclusion}
\label{sec:conclusion}
We proposed the utilization of pseudo anomalies, that are generated using only normal data, to assist the training of an autoencoder (AE) for video anomaly detection. In addition to the conventional training of AEs where a network only attempts to minimize reconstruction error on inputs, we further encourage the network to maximize this loss on pseudo anomalies. The effectiveness of our approach in complementing the AEs to achieve superiority against several existing state-of-the-art models is extensively analyzed on three challenging video anomaly datasets.



\section{Acknowledgements}
This  work  was  supported  by  the  ICT  R\&D  program  of  MSIT/IITP.
[2019-0-01309,  Development  of  AI  Technology for  Guidance  of  a 
Mobile  Robot  to  its  Goal  with  Uncertain  Maps  in  Indoor/Outdoor
Environments].
Also, we thank Ki-In Na and Jae-Yeong Lee for the discussions and support in improving our work.

{\small
\bibliographystyle{ieee_fullname}
\bibliography{egbib}
}

\end{document}